\documentclass[conference,lettersize]{IEEEtran}\IEEEoverridecommandlockouts

\usepackage{multirow}
\usepackage{array}
\usepackage{color}
\usepackage{graphicx}
\ifCLASSINFOpdf
\else
\fi
\usepackage{adjustbox}

\usepackage{color}
\usepackage{url}
\usepackage[utf8]{inputenc}
\begin{document}
\title{ChaLearn Looking at People:\\ A Review of  Events and Resources\thanks{To appear in IJCNN 2017 - IEEE proceedings. Copyright IEEE}}
% \author{\IEEEauthorblockN{
% Sergio Escalera$^{1,2,4}$,
% Hugo Jair Escalante$^{3,4}$,
% Isabelle Guyon$^{4,6}$,
% Xavier Bar\'o$^{2,5}$
% }
% \IEEEauthorblockA{
% $^1$ \emph{Dept. Mathematics and Computer Science, UB}, Spain, \\$^2$ \emph{Computer Vision Center, UAB}, Barcelona, Spain, \\ $^3$ \emph{INAOE}, Puebla, Mexico,  \\$^4$ \emph{ChaLearn}, California, USA, 
%  \\ $^5$ \emph{EIMT/IN3, Open University of Catalonia}, Barcelona, Spain,\\ 
% $^6$ \emph{Universit\'e Paris-Saclay}, Paris, France} 
% }
% \author{\IEEEauthorblockN{Michael Shell}
% \IEEEauthorblockA{School of Electrical and\\Computer Engineering\\
% Georgia Institute of Technology\\
% Atlanta, Georgia 30332--0250\\
% Email: http://www.michaelshell.org/contact.html}
% \and
% \IEEEauthorblockN{Homer Simpson}
% \IEEEauthorblockA{Twentieth Century Fox\\
% Email: homer@thesimpsons.com}
% \and
% \IEEEauthorblockN{James Kirk\\ and Montgomery Scott}
% \IEEEauthorblockA{Starfleet Academy\\
% San Francisco, California 96678--2391\\
% Telephone: (800) 555--1212\\

\author{\IEEEauthorblockN{
Sergio Escalera$^{1,2}$,
Xavier Bar\'o$^{2,3}$,
Hugo Jair Escalante$^{4,5}$,
Isabelle Guyon$^{4,6}$,
}
\IEEEauthorblockA{
$^1$ \emph{Dept. Mathematics and Computer Science, UB}, Spain,\\
$^2$ \emph{Computer Vision Center, UAB}, Barcelona, Spain,\\
$^3$ \emph{EIMT, Open University of Catalonia}, Barcelona, Spain,\\ 
$^4$ \emph{ChaLearn}, California, USA, $^5$ \emph{INAOE}, Puebla, Mexico, 
 \\  $^6$ \emph{Universit\'e Paris-Saclay}, Paris, France, \\
 \tt{sergio.escalera.guerrero@gmail.com}\\
 \tt{http://chalearnlap.cvc.uab.es}}}


%Telephone: (800) 555--1212, Fax: (888) 555--1212}


\maketitle

%% For IJCNN, Hugo is leading the writing of a paper summarizing the next competition (without baselines, but required to assure the success of IJCNN special session and competition). We all will be involved in this publication. Hugo, you can use a lot of information from CVPR proposal where we are updating new information until deadline next monday: https://docs.google.com/document/d/1WWuL-DctvbLRcZNgQWYVWDVe032z2rLnlhY5wt2mSfw/edit

\begin{abstract}
This paper reviews the historic of \emph{ChaLearn Looking at People} (LAP) events. We started in 2011 (with the release of the first Kinect device) to run challenges related to human action/activity and gesture recognition.  Since then we have regularly organized events in a series of competitions covering all aspects of visual analysis of humans.  So far we have organized  more than 10 international challenges and events in this field. This paper reviews associated events, and introduces the ChaLearn LAP platform where public resources (including code, data and preprints of papers) related to the organized events are available. We also provide a discussion on our main findings and perspectives of ChaLearn LAP activities. 
%we make public several resources from the organized events, and finally discuss about future plans.
\end{abstract}

% \fi
\IEEEpeerreviewmaketitle

\section{Introduction}

Looking at People (LAP) is a challenging area of research that deals with the problem of automatically recognizing people in images, detecting and describing body parts, inferring their spatial configuration, performing action/gesture recognition from still images or image sequences, often including multi-modal data. Any scenario where the visual or multi-modal analysis of people takes a key role is of interest  within the, so called,  field of Looking at People~\cite{0}.

Because of the huge configuration space of human bodily structure, posture, and movement, human analysis is a difficult problem for Computer Vision and Machine Learning, which involves dealing with numerous distortion factors, including: illumination changes, partial occlusions, changes in the point of view, rigid and elastic deformations, or high inter and intra-class variability. Despite the high difficulty of the problem, modern Computer Vision and Machine Learning techniques have advanced greatly the state-of-the-art and therefore deserve further attention.

Our selection of tasks when organizing new events is motivated both by academic interest and by potential applications, such as TV production, home entertainment (multimedia content analysis), education purposes, sociology research, surveillance and security, improved quality live by means of monitoring or automatic artificial assistance, etc. Academic interest in illustrated by new burgeoning subareas of LAP, such as Affective Computing, Social Signal Processing, Human Behavior Analysis, and Social Robotics, providing us with opportunities to make connections with other domains and expose the Computer Vision and Machine Learning community to new problems of high practical and societal interest.

These objectives were first illustrated in 2011 when ChaLearn LAP organized its first event on one-shot multi-modal gesture recognition from RGB-Depth data sources. Since then, ChaLearn LAP has organized over 10 international challenges in the field of LAP, including face analysis, body posture, action/gesture analysis, including still images, image sequences, and multi-modal data sources, and focusing in cutting edge trends in computer vision and pattern recognition, like explainable computer vision, personality analysis, and collaborative competitions (a.k.a. \emph{coopetitions}).

In this paper we briefly review  the  events organized so far, which are not limited to challenges, but also include, workshops, special issues, and book series. We also introduce the new ChaLearn LAP platform, which contains all information and resources from previous and current events, including programs, papers, codes, and data, among others. Finally, we discuss about near future plans within ChaLearn LAP series. 

\section{ChaLearn LAP Events}
\label{sec:rw}

%ChaLearn~\footnote{\url{http://chalearn.org}} is a non-profit organization   with extensive experience in the organization of academic challenges in machine learning and related fields. 
ChaLearn \url{http://chalearn.org} is a non-profit organization with vast experience in the organization of academic challenges. ChaLearn is interested in all aspects of challenge organization, including data gathering procedures, evaluation protocols, novel challenge scenarios (e.g., coopetitions), training for challenge organizers, challenge analytics, results dissemination and, ultimately, advancing the state-of-the-art through challenges. %As a sample we provide a summary of recent successful challenges:
% in the fields of machine learning and computer vision. 
Since 2003, ChaLearn has been organizing a number of challenges in several interrelated fields, including machine learning, computer vision, pattern recognition, causality and bioinformatics.  The first challenge we organized was the NIPS 2003 Feature Selection challenge\footnote{\url{http://clopinet.com/isabelle/Projects/NIPS2003/}}, and currently several challenges are running (please visit ChaLearn's website for update information on ongoing and past challenges~\url{http://chalearn.org}). 
%ChaLearn is a non-profit organization with extensive experience in the organization of academic challenges in the fields of machine learning and computer vision. Our challenge topics are selected because of their high impact and timeliness and are generally organized in conjunction with workshops in major machine learning or computer vision conferences (NIPS, ICML, CVPR, ICCV, etc.),

ChaLearn LAP is a division of ChaLearn that focuses in challenges in the fields of computer vision and pattern recognition.  This section  reviews the historic of ChaLearn LAP events. We split them into challenges and their associated workshops, special issues, and Challenges in Machine Learning (CIML) Springer book series. Table~\ref{tablesummary} summarizes the list of challenges and their associated workshops and special issues. Next, we briefly review the organized competitions in the period 2011-2016. %The summary of events and best results for each of the organized competitions are shown in Table~\ref{tablesummary}.

\subsection*{Challenges and Workshops}
Kinect revolutionized in 2010 the field of gesture recognition, and had a broad impact into other LAP areas.  It i a valuable resource as it provides a variety of data modalities, including RGB image, depth image (using an infrared sensor), and audio. For this reason the first three challenges organized by ChaLearn LAP relied on the use and exploitation of such rich information.

\subsubsection*{2011/12 Gesture Challenge}

First, we organized a challenge on gesture and sign language recognition from video, mostly focusing on hand and arm gestures, although facial expressions and whole body motion may enter into account~\cite{36}, with funding from DARPA, NSF, and the Pascal2 EU network of excellence, and prizes donated by Microsoft Xbox and Texas Instrument. We used the Kaggle platform. Applications include recognizing signals for man-machine communication, translating sign languages for the deaf to hearing people, and computer gaming. The challenge focused in one-shot gesture recognition (i.e., learning models from a single example), and included qualitative and quantitative tracks. A large ``user-independent" dataset (of over 60'000 gestures) covering a high number of domains and gesture categories was released~\cite{CGD}, and the Leveinstein metric was used as a quantitative measure for ranking participants.  Dozens of participants joined the competition that lasted more than one year. Although the impulse for the design of this challenge and part of the funding was given by the DARPA Deep Learning program, {\em no deep learning method} was applied! Rather, more conventional combinations of ad hoc feature extraction methods and hidden Markov models dominated the methodology of top ranking participants. The {\em no deep learning revolution} caught up right after this first challenge!

\subsubsection*{2013 Looking at People ICMI Challenge}
ChaLearn LAP organized in 2013 a challenge and workshop on multi-modal gesture recognition from 2D and 3D video data using Kinect~\cite{37}. In this challenge and all following ones, we used the Codalab platform running on Microsoft Azure, with the support of Microsoft Research. New in this competition was the user independent aspect capturing applications of gesture recognition genuinely important in many multi-modal interaction and computer vision applications, including image/video indexing, video surveillance, computer interfaces, and gaming. User-independent recognition of continuous, natural signing is very challenging due to the multimodal nature of the visual cues (e.g., movements of fingers and lips, facial expressions, body pose), as well as technical limitations such as spatial and temporal resolution and unreliable depth cues. We ran a competition containing a dataset with more than 13'000 gesture samples from a dictionary of 20 Italian sign gesture categories. For the first time in video LAP challenges deep learning methods won! An interesting finding from this challenge was that the audio modality was very helpful for recognizing gestures. 

\subsubsection*{2014 Looking at People ECCV Challenge}
ChaLearn organized in 2014 three parallel challenge tracks on Human Pose Recovery on RGB data, action/interaction spotting on RGB data, and gesture spotting on RGB-Depth data~\cite{38}.\\
The challenge featured three quantitative tracks:\\
\begin{itemize}
\item Track 1: Human Pose Recovery. More than 8,000 frames of continuous RGB sequences are recorded and labeled with the objective of performing human pose recovery by means of recognizing more than 120'000 human limbs of different people.
\item Track 2: Action/Interaction Recognition. 235 performances of 11 action/interaction categories are recorded and manually labeled in continuous RGB sequences of different people performing natural isolated and collaborative actions randomly.
\item Track 3: Gesture Recognition. More than 14'000 gestures are drawn from a vocabulary of 20 Italian sign gesture categories. The emphasis of this third track is on multi-modal automatic learning of a set of 20 gestures performed by several different users, with the aim of performing user independent continuous gesture spotting.
\end{itemize}

During this challenge, the state of the art was significantly advanced in the three tracks, where solutions based in deep learning obtained the best results. 

\subsubsection*{2015 Looking at People CVPR Challenge}
After the boom of Kinect and the fact that human-level performance was reached in several tasks/challenges involving Kinect-recorded data, ChaLearn LAP started to look at open challenges in relevant fields that have to do with still and sequences of RGB images. 

ChaLearn organized in 2015 parallel challenge tracks on RGB data for Human Pose Recovery, action/interaction spotting, and cultural event recognition~\cite{40}. The challenge featured three quantitative tracks:
\begin{itemize}
\item Track 1: Human Pose Recovery. More than 8,000 frames of continuous RGB sequences are recorded and labeled with the objective of performing human pose recovery by means of recognizing more than 120,000 human limbs of different people.
\item Track 2: Action/Interaction Recognition. 235 performances of 11 action/interaction categories are recorded and manually labeled in continuous RGB sequences of different people performing natural isolated and collaborative actions randomly.
\item Track 3: Cultural Event Recognition. More than 10'000 images corresponding to 50 different cultural event categories will be considered. In all the categories, garments, human poses, objects and context will be possible cues to be exploited for recognizing the events, while preserving the inherent inter- and intra-class variability of this type of images. Examples of cultural events will be Carnival, Oktoberfest, San Fermin, Maha-Kumbh-Mela and Aoi-Matsuri, among others.
\end{itemize}

\subsubsection*{2015 Looking at People ICCV Challenge}
ChaLearn organized for ICCV2015 two parallel quantitative challenge tracks on RGB data~\cite{39}.
\begin{itemize}
\item Track 1: Apparent Age Estimation. 5'000 images each displaying a single individual, labeled with the apparent age. Each image has been labeled by multiple individuals using a collaborative Facebook implementation. The votes variance is used as a measure of the error for the predictions. This is the first state of the art dataset for Apparent Age Recognition rather than Real Age recognition.
\item Track 2: Cultural Event Recognition. Near 30'000 images corresponding to 100 different cultural event categories will be considered. In all the categories, garments, human poses, objects and context will be possible cues to be exploited for recognizing the events, while preserving the inherent inter- and intra-class variability of this type of images. %Examples of cultural events are Carnival, Oktoberfest, San Fermin, Maha-Kumbh-Mela, Aoi-Matsuri. 
\end{itemize}

\subsubsection*{2016 Looking at People CVPR Challenge}
ChaLearn organized for CVPR2016 three parallel quantitative challenge tracks on RGB face analysis~\cite{41}.
\begin{itemize}
\item Track 1: Apparent Age Estimation. An extended version of the previous ICCV2015 challenge dataset. It contains 8,000 images each displaying a single individual, labeled with the apparent age. Each image has been labeled by multiple individuals, using a collaborative Facebook implementation and Amazon Mechanical Turk. The votes variance is used as a measure of the error for the predictions. This is the first state of the art dataset for Apparent Age Recognition rather than Real Age recognition.
\item Track 2: Accessories Classification. The aim of this track is to detect and classify complements and accessories worn by the subjects. It uses a fraction of the Faces of the World dataset, a challenging dataset consiting of 8'000 images, each displaying a single individual, labelled with the accessories they are wearing. 
\item Track 3: Smile and Gender Classification. In this track, participants will have to classify images of the FotW dataset according to gender (male, female or other) and basic expression (smiling, neutral or other expression). Along with accurate face detection and aligment, this track will require robust feature selection and extraction in order to identify the subject’s gender and expression, which can be difficult to classify even to the human eye in uncontrolled environments such as those present in the FotW dataset.
\end{itemize}

\subsubsection*{2016 Looking at People ECCV Challenge}
Research advances in computer vision and pattern recognition have resulted in tremendous advances in different problems and applications. As a result, several problems on visual analysis can be considered as solved (e.g., face recognition), at least in certain scenarios and under specific circumstances.  Despite these important advances, there are still many open problems that are receiving much attention from the community because of the potential applications. For that reason, ChaLearn LAP has started to focus on problems going beyond the information visually depicted in videos. 

First, co-located with the workshop we organized a challenge on ``first impressions", in which participants developed solutions for recognizing personality traits of users in short video sequences~\cite{42}. We made available a large newly collected dataset sponsored by Microsoft of at least 10'000 15-second videos collected from YouTube, annotated with personality traits by AMT workers. The traits correspond to the ``big five" personality traits used in psychology and well known of hiring managers using standardized personality profiling: Extroversion, agreeableness, conscientiousness, neuroticism, and openness to experience. As is known, the first impression made is highly important in many contexts, such as human resourcing or job interviews. This work could become very relevant to training young people to present themselves better by changing their behavior in simple ways. 

\subsubsection*{2016 Looking at People ICPR Challenge}
In the same line, we organized a contest around four LAP problems that require, in addition to performing an effective visual analysis, to deal with multimodal information (e.g., audio,  RGB-D video, etc.) in order to be solved~\cite{43}. These four tracks included  a second round on the first impressions challenge.  %In addition we focused on problems in which we aim to recognize non visually evident patterns (e.g., personality traits). 
The contest was supported by three organizations with vast experience and prestige in the organization of academic contests, namely: Chalearn, MediaEval and ImageCLEF. The contest was also supported by the IAPR TC 12 on visual and multimedia information systems.\\
The tracks of the contest were as follows:
\begin{itemize}
\item Track 1: First impressions challenge. participants developed solutions for recognizing personality traits of users in short video sequences. This corresponds to the second round of the first impressions challenge organized at ECCV 2016~\cite{42}.
\item Track 2: Isolated gesture recognition track. We organized a track on isolated gesture recognition from RGB-D data, where the goal was to develop methods for recognizing the category of human gestures from segmented RGB-D video. A new dataset called ChaLearn LAP RGB-D Isolated Gesture Dataset (IsoGD) was considered for this track~\cite{isoGD}.  This dataset includes 47'933 RGB-D gesture videos (about 9G). Each RGB-D video depicts a single  gesture and there are 249 gesture categories performed by 21 different individuals. 

\item Track 3: Continuous gesture recognition track. We also organized a track complimentary to track 2 on continuous gesture recognition from RGB-D data. The goal in this track was to develop methods that can perform simultaneous segmentation and recognition of gesture categories from continuous RGB-D video. The newly created:  ChaLearn LAP RGB-D Continuous Gesture Dataset (ConGD) is considered for this track~\cite{isoGD}. The dataset comprises a total of 47,933 gestures in 22535 RGB-D continuous videoss (about 4G).
Each RGB-D video depicts one or more gestures  and there are 249 gesture categories performed by 21 different individuals.

\item Track 4: Context of experience track. The aim was to explore the suitability of video content for watching in certain situations. Specifically, we look at the situation of watching movies on an airplane. As a viewing context, airplanes are characterized by small screens and distracting viewing conditions. We assume that movies have properties that make them more or less suitable to this context. We were interested in developing systems that are able to reproduce a general judgment of viewers about whether a given movie is a good movie to watch during a flight. We provided a dataset including a list of movies and human judgments concerning their suitability for airplanes~\cite{CoEdata}. The goal of the task was to use movie metadata and audio-visual features extracted from movie trailers in order to automatically reproduce these judgments. The provided dataset comprised: a total of 318 movies + metadata + audio (MFCC), textual (td-idf) and visual (HOG, CM, LBP, GLRLM) features + movie. Video resolution: full HD 1080p, with length of around 2-5 minutes, the considered categories are: latest, recent, the collection, family, world, dutch and European. 
\end{itemize}

\subsection*{Special issues}

Together with the challenges and their associated workshops, we organized a series of Special Issues and Special Topics related to LAP in prestigious  journals. In the following we provide a summary of each of them.

\subsubsection*{2014 JMLR Gesture}
We organized a Special Topic on Gesture Recognition at Journal of Machine Learning Research. The scope of this special issue included: Algorithms for gesture and activity recognition, in particular addressing: Learning from unlabeled or partially labeled data; Learning from few examples per class, and transfer learning; Continuous gesture recognition and segmentation Deep learning architectures, including convolutional neural networks; Gesture recognition in challenging scenes, including cluttered/moving backgrounds or cameras; Large scale gesture recognition o Multi-modal features for gesture recognition, including non-conventional input sources, such as inertial, depth or thermal data; Integrating information from multiple channels (e.g., position/motion of multiple body parts, hand shape, facial expressions); Applications in video surveillance , image or video indexing and retrieval, recognition of sign languages for the deaf, emotion recognition and affective computing, computer interfaces, virtual reality, robotics, ambient intelligence, and games. A total of 18 papers were published within this Special Topic at Journal of Machine Learning Research~\cite{1,2,3,4,5,6,7,8,9,10,11,12,13,14,15,16,17,18}

\subsubsection*{2016 TPAMI HuPBA}
We organized a Special Issue on Multi-modal Human Pose Recovery and Behavior Analysis (HuPBA) at IEEE Transactions on Pattern Analysis and Machine Intelligence Journal. The scope of this special issue included: Multi-modal data fusion and learning strategies; Combination techniques of visual and non-visual descriptors (RGB+D, multispectral, thermal, IR, audio, IMU, EDA, EGG, etc.); Calibration and synchronization of multi-modal data; Multi-modal datasets and evaluation metrics; Leisure, security, health and energy applications based on multi-modal data; Multi-modal Affective Computing and Social Signal processing systems; Multi-modal algorithms designed for GPU, smart phones and game consoles. A total of 17 papers were published withiin this Special Issue at IEEE TPAMI~\cite{19,20,21,22,23,24,25,26,27,28,29,30,31,32,33,34,35}.

\subsubsection*{2016 IJCV LAP}
We organized a Special Issue on LAP at International Journal of Computer Vision. The scope of this special issue included: Gesture (hands and body), posture and sign recognition, analysis and synthesis; Face recognition, analysis and synthesis; Body motion analysis and synthesis, action/interaction recognition and spotting; Context analysis for human behavior recognition in still images and image sequences; Psychological, affective, and behavioral analysis; Tracking systems on Looking at People; 3D human analysis and multi-modal Looking at People; Biometry analysis, identification and verification; Looking at People for Virtual and Augmented Reality; Datasets and evaluation protocols on Looking at People; Computer Vision applications of LAP. 

\subsubsection*{2017 TPAMI Faces}
Currently we are organizing a Special Issue at IEEE Transactions on Pattern Analysis and Machine Intelligence journal in the topic of face analysis. Automated face analysis is a research topic that has received so much attention from the Computer Vision and Pattern Recognition communities in the past, that research progress has made some think that problems such as face recognition or face detection are solved. However, many aspects of face analysis remain open problems, including the implementation of large-scale face detection and recognition methods for images captured in real-life applications. Thse include 3D face analysis and face pose estimation from 2D; face analysis under extreme pose variation and occlusion, identity recognition, emotion and micro-expression recognition, and analysis of dynamics of facial expression. Luckily, these are also areas in which the community is making rapid progress enabled by more powerfull methods (e.g. Deep Learning) that push the state-of-the-art. Among real-world applications are security and video surveillance, human computer/robot interaction, interpersonal communication, entertainment, commerce, and assistive technologies for education and physical and mental health.

\subsubsection*{2017 TAC Personality}
Currently we are organizing a Special Issue at IEEE Transactions on Affective Computing in the topic of personality analysis. The automatic analysis of videos and any other kind of input data to characterize human behavior has become an area of active research with applications in affective computing, human-machine interfaces, gaming, security, marketing, health, and other domains. Research advances in multimedia information processing, computer vision and pattern recognition have lead to established methodologies that are able to successfully recognize consciously executed actions, or intended movements (e.g., gestures, actions, interactions with objects and other people). However, recently there has been much progress in terms of computational approaches to characterize sub-conscious behaviors, which may be revealing aptitudes or competence, hidden intentions, and personality traits. This special issue will compile progress on apparent personality analysis from a computational perspective.

\subsection*{CIML Series}

Starting in 2017, we are running the Springer Series on Challenges in Machine Learning\url{http://www.springer.com/series/15602}, the first series of books entirely dedicated to collect papers associated to  successful competitions in machine learning and related fields.  The scope of the series also includes analyses of the challenges, tutorial material, dataset descriptions, and pointers to data and software. Together with the websites of the challenge competitions, they offer a complete teaching toolkit and a valuable resource for engineers and scientists. This new series will be one of our main forums for disseminating results of competitions in Chalearn LAP.

\subsection*{Discussion}

We just described the challenges organized by Chalearn LAP in the last 6 years. The areas covered by the different challenges, included the analysis of human faces, body movements, actions, pose estimation and even analysis of cultural images.  Clearly, the organization of these events helped considerably to advance the state of the art in different subfields of computer vision and pattern recognition. Multiple information sources were considered through the organized competitions, including RGB images and video, depth video and audio. Approaches taking advantage of multimodal information were the most successful in most of the cases.   Interestingly, with the organization of the different challenges we witnessed the empowerment and the establishment of deep learning as the methodology ruling computer vision competitions. We foresee upcoming challenges will be dominated by deep learning as well, however,  a major drawback of this sort of methods is that they are limited in terms of their explicability and interpretability features. Because of that, we are currently organizing a challenge in that direction~\cite{225}. We foresee that challenges in similar topics  (i.e., extending the capabilities of deep learning techniques) will prevail the arenas of computer vision and patter recognition.  

We also reviewed and presented an overview of the forums and publications at which challenge results have been presented, including workshops, special issues and the brand new CIML Series. This is an aspect as important as the challenges themselves, advancing the state of the art through competitions require of effective dissemination means. In the near future we will focus on novel ways of challenge dissemination.

%\cleardoublepage

%\subsection*{Summary table}
\begin{table*}[ht]
\centering
\begin{tabular}{|p{3.1cm}|p{3.1cm}|p{3.1cm}|p{3.1cm}|p{3.1cm}|}
\hline\vspace{+0.1cm}
\textbf{Tracks} & \vspace{+0.1cm}\textbf{Workshops} & \vspace{+0.1cm} \textbf{Dataset} & \vspace{+0.1cm} \textbf{Special issue} & \vspace{+0.1cm} \textbf{Winner and score}\vspace{+0.1cm} \\ \hline\hline
\vspace{+0.005cm}   One-Shot-Learning Gesture Challenge & Gesture recognition workshops CVPR 2011/2012, ICPR 2012 & CGD 2011 Data & IJCV LAP 2016 & alfnie: 0.07099 (Error) \\ \hline
\vspace{+0.005cm} Multimodal Gesture Recognition & Multi-modal Gesture Recognition Workshop ICMI 2013 & Multimodal Gesture Recognition: Montalbano V1 & JMLR Gesture 2014, TPAMI HuPBA 2016 and IJCV LAP 2016 & IVA MM: 0.12756 (Error) \cite{44} \\ \hline
\vspace{+0.005cm} Human Pose Recovery (First round) & ChaLearn LAP Workshop ECCV 2014 & Human Pose & TPAMI HuPBA 2016 and IJCV LAP 2016 & ZJU: 0.194144 (Accuracy) \cite{45} \\ \hline
\vspace{+0.005cm} Action/Interaction Recognition (First round) & ChaLearn LAP Workshop ECCV 2014 & Action/Interaction Recognition & JMLR Gesture 2014, TPAMI HuPBA 2016 and IJCV LAP 2016 & CUHK-SWJTU: 0.507173 (Accuracy) \cite{46} \\ \hline
\vspace{+0.005cm} Gesture Recognition & ChaLearn LAP Workshop ECCV 2014 & Multimodal Gesture Recognition: Montalbano V2 & JMLR Gesture 2014, TPAMI HuPBA 2016 and IJCV LAP 2016 & LIRIS: 0.849987 (Accuracy) \cite{47} \\ \hline
\vspace{+0.005cm} Action/Interaction Recognition (Second round) & ChaLearn LAP Workshop CVPR 2015 & Action/Interaction Recognition & JMLR Gesture 2014, TPAMI HuPBA 2016 and IJCV LAP 2016 & MMLAB: 0.855 (Accuracy) \cite{48} \\ \hline
\vspace{+0.005cm} Cultural Event Recognition (First round) & ChaLearn LAP Workshop CVPR 2015 & Cultural Event V1 & IJCV LAP 2016 & VIPL-ICT-CAS: 0.854 (Accuracy) \cite{49} \\ \hline
\vspace{+0.005cm} Apparent age Estimation (First round) & ChaLearn LAP Workshop ICCV 2015 & Apparent age V1 & IJCV LAP 2016 and TPAMI Faces 2017 & CVL ETHZ: 0.264975 (Error) \cite{50} \\ \hline
\vspace{+0.005cm} Cultural Event Recognition (Second round) & ChaLearn LAP Workshop ICCV 2015 & Cultural Event V2 & IJCV LAP 2016 & SIAT MMLAB: 0.9349 (Accuracy) \cite{51} \\ \hline
\vspace{+0.005cm} Apparent age Estimation (Second round) & ChaLearn LAP Workshop CVPR 2016 & Apparent age V2 & IJCV LAP 2016 and TPAMI Faces 2017 & OrangeLabs: 0.2411 (Error) \cite{52} \\ \hline
\vspace{+0.005cm} Accessories Classification & ChaLearn LAP Workshop CVPR 2016 & Accessories Classification & IJCV LAP 2016 & SIAT MMLAB:	0.9349 (Accuracy) \cite{53} \\ \hline
\vspace{+0.005cm} Smile and Gender Classification & ChaLearn LAP Workshop CVPR 2016 & Smile and Gender Classification & IJCV LAP 2016 & SIAT MMLAB: 0.8926 (Accuracy) \cite{54} \\ \hline
\vspace{+0.005cm} First impressions Challenge (First round) & ChaLearn LAP Workshop ECCV 2016 & First impressions & IJCV LAP 2016, TAC Personality 2017 and TPAMI Faces 2017 & NJU-LAMDA: 0.4025 (Accuracy) \cite{55} \\ \hline
\vspace{+0.005cm} First impressions Challenge (Second round) & ChaLearn LAP Workshop ICPR 2016 & First impressions & IJCV LAP 2016, TAC Personality 2017 and TPAMI Faces 2017 & BU-NKU: 0.913 (Accuracy) \cite{56} \\ \hline
\vspace{+0.005cm} Isolated Gesture Recognition & ChaLearn LAP Workshop ICPR 2016 & Isolated Gesture Recognition & JMLR Gesture 2014, TPAMI HuPBA 2016 and IJCV LAP 2016 & FLiXT: 0.569 (Accuracy) \cite{57} \\ \hline
\vspace{+0.005cm} Continuous Gesture Recognition & ChaLearn LAP Workshop ICPR 2016 & Continuous Gesture Recognition & JMLR Gesture 2014, TPAMI HuPBA 2016 and IJCV LAP 2016 & ICT NHCI: 0.2869 (Accuracy) \cite{58} \\ \hline
\vspace{+0.005cm} Context of Experience Track & ChaLearn LAP Workshop ICPR 2016 & Context of Experience & IJCV LAP 2016 & uklu: 0.69697 (Accuracy) \\ \hline
\end{tabular}\vspace{+0.1cm}\centering\caption{Summary of organized ChaLearn LAP events and results (2011-2017).}\label{tablesummary}
\end{table*}

\section{ChaLearn LAP webpage}

ChaLearn LAP official webpage can be found at \url{http://chalearnlap.cvc.uab.es/}. It contains all material associated to the organized events, including challenges, competitions, data, codes, special issues, books, and associated publications. In the Challenges menu all organized and in progress competitions are listed. Subpages from competitions include:
\begin{itemize}
\item Description: Full explanation of the competition.
\item Schedule: Important dates of the competition.
\item Associated events: Links to related events.
\item Tracks: The set of tracks within the same competition. Each one includes the results of the participants.
\item People: It shows the list of co-organizers. 
\item Sponsors: The set of sponsors of the competition.
\end{itemize}
Datasets menu lists all available datasets. Subpages from datasets include:
\begin{itemize}
\item Description: Full explanation of the dataset.
\item Associated events: Links to related events.
\item Data: Training, Validation and Test data. 
\item Results: The results of track participants. Users can also upload their results post competition. It also contains the summary of methods and open source code. 
\end{itemize}
Workshop menu lists all organized workshops.
\begin{center} \includegraphics[scale=0.20]{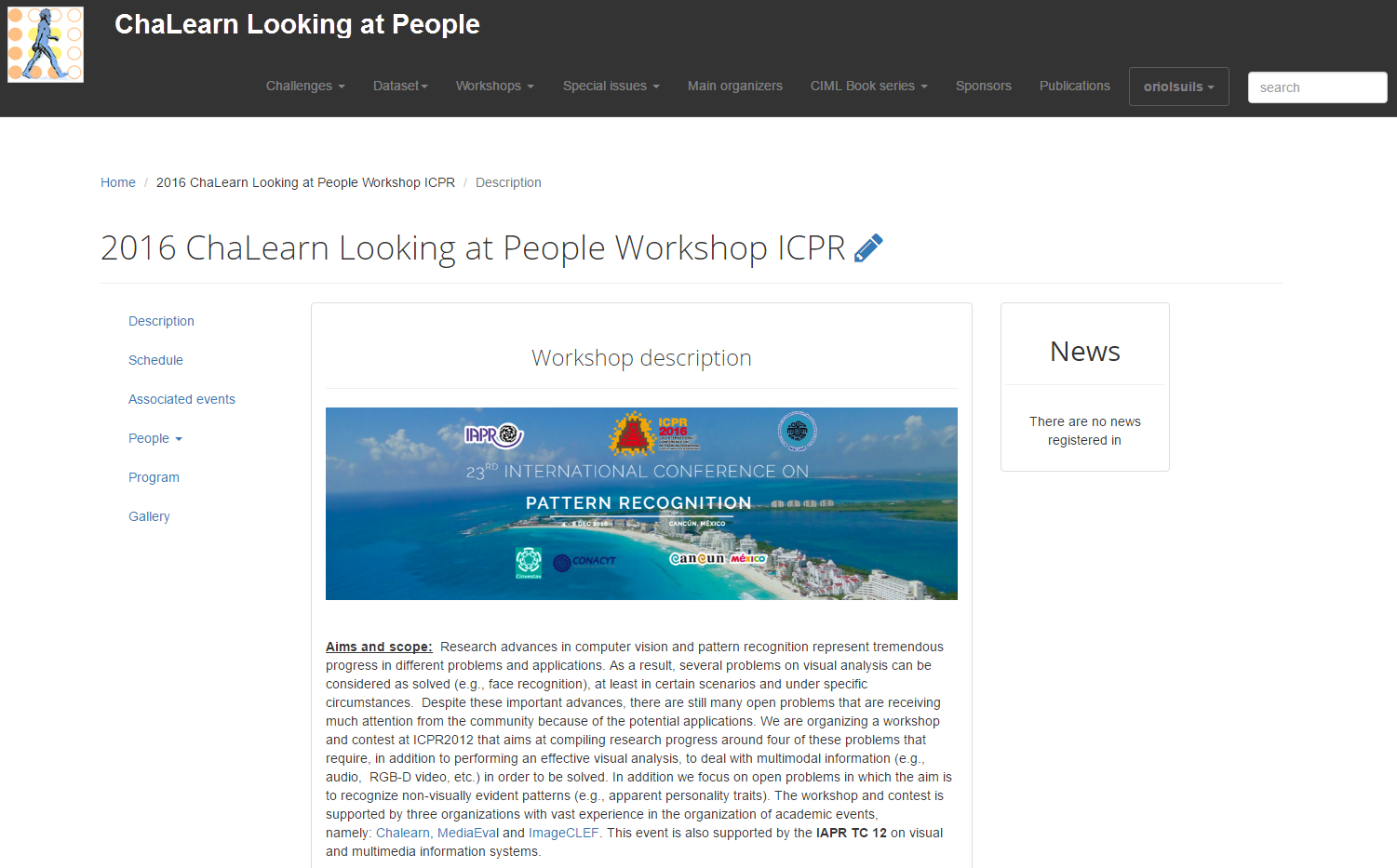} \end{center}
Subpages of workshop page include:
\begin{itemize}
\item Description: Full explanation of the workshop.
\item Schedule: Important dates of the workshop.
\item Associated events: Links to related events.
\item People: List of co-organizers.
\item Program: Calendar of the workshop day.
\item Gallery: Set of images of the workshop day.
\end{itemize}
The special issues menu list all organizers special issues. Subpages from special issues include:
\begin{itemize}
\item Description: Full explanation of the special issue.
\item Schedule: Important dates of the special issue.
\item Associated events: Links to related events.
\item People: List of guest editors.
\item Papers published: Set of papers published within the special issue.
\end{itemize}
Last web pages contain the main ChaLearn LAP organizers, general sponsors and set publications related to ChaLearn LAP events.

\section{Final remarks}

We reviewed the historic of events organized within the ChaLearn LAP series, including the competitions, workshops, special issues and books in the period 2011-2017. We also presented the new ChaLearn LAP webpage, which includes all the information associated to the organizing events, also including data, code, and the results of the competitions and the state of the art results on the provided datasets. Users of the webpage can also submit predictions to any data used within ChaLearn LAP competitions. Currently, we are organizing a new series of competitions within the field of LAP, including face and body analysis. In particular, 2017 events will focus on explainable computer vision and machine learning models within the field of personality analysis.

\section*{Acknowledgments}

We thank Oriol Suils for his collaboration in the implementation of the new ChaLearn LAP webpage. We thank all the participants and co-organizers of the organized ChaLearn LAP events. We appreciate the support of the sponsors of the different organized events, including Microsoft Research, Google, Nvidia Corporation, Amazon, Facebook, and Disney Research, among others. This work has been partially supported by Spanish projects TIN2013-43478-P, TIN2015-66951-C2-2-R and TIN2016-74946-P,  the European Commission Horizon 2020 granted project SEE.4C under call H2020-ICT-2015 and CONACYT under project grant  CB2014-241306.  This effort was initiated by the DARPA Deep Learning program and was supported by the US National Science Foundation (NSF) under grants ECCS 1128436 and ECCS 1128296, the EU Pascal2 network of excellence and the Challenges in Machine Learning organization (ChaLearn). Any opinions, findings, and conclusions or recommendations expressed in this material are those of the authors and do not necessarily reflect the views of the sponsors.



%\bibliographystyle{IEEEtran}
%\bibliography{IEEEabrv,bibliog}

\begin{thebibliography}{50}
	\bibitem{0}Thomas B. Moeslund, Adrian Hilton, Volker Krüger, Leonid Sigal (Editors). Visual Analysis of Humans: Looking at People.  Springer, 2011 edition.
    \bibitem{36}Isabelle Guyon, V. Athitsos, P. Jangyodsuk, H. J. Escalante, B. Hamner, Advances in Depth Image Analysis and Applications, Lecture Notes in Computer Science Volume 7854, Springer 2013, pp 186-204.
  \bibitem{CGD} Isabelle Guyon, Vassilis Athitsos, Pat Jangyodsuk, Hugo Jair Escalante. The ChaLearn Gesture Dataset CGD 2011, Machine Vision and Applications, (2014) Vol. 25(8):1929--1951.      
  \bibitem{37}Sergio Escalera, Jordi Gonzàlez, Xavier Baró, Miguel Reyes, Oscar Lopés, Isabelle Guyon, Vassilis Athitsos, Hugo J. Escalante, Multi-modal Gesture Recognition Challenge 2013: Dataset and Results, Chalearn Multi-Modal Gesture Recognition Workshop, International Conference on Multimodal Interaction, ICMI, 2013.
\bibitem{38}Sergio Escalera, Xavier Baró, Jordi Gonzàlez, Miguel A. Bautista, Meysam Madadi, Miguel Reyes, Víctor Ponce-López, Hugo J. Escalante, Jamie Shotton, Isabelle Guyon, ChaLearn Looking at People Challenge 2014: Dataset and Results, ECCV, 2014.
\bibitem{40}Xavier Baro, Jordi Gonzalez, Junior Fabian, Miguel A. Bautista, Marc Oliu, Hugo J. Escalante, Isabelle Guyon, Sergio Escalera, ChaLearn Looking at People 2015 challenges: action spotting and cultural event recognition, CVPR workshops, 2015.
\bibitem{39}Sergio Escalera, Junior Fabian, Pablo Pardo, Xavier Baró, Jordi Gonzàlez, Hugo J. Escalante, Dusan Misevic, Ulrich Steiner, Isabelle Guyon, ChaLearn Looking at People 2015: Apparent Age and Cultural Event Recognition datasets and results, ICCV, 2015
\bibitem{41}Sergio Escalera, Mercedes Torres Torres, Brais Martínez, Xavier Baró, Hugo Jair Escalante, Isabelle Guyon, Georgios Tzimiropoulos, Ciprian Corneanu, Marc Oliu, Mohammad Ali Bagheri, Michel Valstar, ChaLearn Looking at People and Faces of the World: Face Analysis Workshop and Challenge 2016, CVPR, 2016
\bibitem{42}Víctor Ponce-López, Baiyu Chen, Marc Oliu, Ciprian Cornearu, Albert Clapés, Isabelle Guyon, Xavier Baró, Hugo Jair Escalante, Sergio Escalera, ChaLearn LAP 2016: First Round Challenge on First Impressions - Dataset and Results, ECCV, 2016
\bibitem{43}Hugo Jair Escalante, Víctor Ponce-López, Jun Wan, Michael A. Riegler, Baiyu Chen11, Albert Clapés, Sergio Escalera, Isabelle Guyon, Xavier Baró, Pål Halvorsen, Henning Müller and Martha Larson, ChaLearn Joint Contest on Multimedia Challenges Beyond Visual Analysis: An overview, ICPR, 2016
\bibitem{isoGD} Jun Wan, Yibing Zhao, Shuai Zhou,  Isabelle Guyon, and Sergio Escalera and Stan Z. Li, "ChaLearn Looking at People RGB-D Isolated and Continuous Datasets for Gesture Recognition", CVPR workshop, 2016.
\bibitem{CoEdata} M. Riegler, M. Larson, C. Spampinato, P. Halvorsen, M. Lux, J. Markussen, K. Pogorelov, C. Griwodz, and H. Stensland. Right insight: A dataset for exploring the automatic prediction of movies suitable for a watching situation, in Proc. of 7th International Conference on Multimedia Systems. ACM, 2016, pp. 45:1–45:6 



	\bibitem{1}Yang Wang, Duan Tran, Zicheng Liao, and David Forsyth, Discriminative Hierarchical Part-based Models for Human Parsing and Action Recognition, Journal of Machine Learning Research 13 (2012) 3075-3102.
    \bibitem{2}Thad Starner, MAGIC Summoning: Towards automatic suggesting and testing of gestures with low probability of false positives during use, Journal of Machine Learning Research 14 (2013) 209-242.
	\bibitem{3}Aleix Martinez and Shichuan Du, A Model of the Perception of Facial Expressions of Emotion by Humans: Research overview and perspectives, Journal of Machine Learning Research 13 (2012) 1589-1608.
	\bibitem{4}Helen Cooper, Richard Bowden, Eng Jon Ong, and Nicolas Pugeault, Sign Language Recognition using Sub-Units, Journal of Machine Learning Research 13 (2012) 2205-2231.
\bibitem{5}Kester Duncan, Finding Recurrent Patterns from Continuous Sign Language Sentences for Automated Extraction of Signs, Journal of Machine Learning Research 13 (2012) 2589-2615.
\bibitem{6}Yui Man Lui, Human Gesture Recognition on Product Manifolds, Journal of Machine Learning Research 13 (2012) 3297-3321.
\bibitem{7}Anastasios Roussos, Stavros Theodorakis, Vassilis Pitsikalis, and Petros Maragos, Dynamic Affine-Invariant Shape-Appearance Handshape Features and Classification in Sign Language Videos, Journal of Machine Learning Research 14 (2013) 1627-1663.
\bibitem{8}Norberto Goussies, Sebastián Ubalde, and Marta Mejail, Transfer Learning Decision Forests for Gesture Recognition, Journal of Machine Learning Research 15 (2014) 3847-3870.
\bibitem{9}Ifeoma Nwogu and Manavender Malgireddy, Language-Motivated Approaches to Action Recognition, Journal of Machine Learning Research 14 (2013) 2189-2212.
\bibitem{10}Sean Ryan Fanello and Ilaria Gori, Keep It Simple and Sparse: Real-Time Action Recognition, Journal of Machine Learning Research 14 (2013) 2617-2640.
\bibitem{11}Jun Wan, One-shot Learning Gesture Recognition from RGB-D Data Using Bag-of-Features, Journal of Machine Learning Research 14 (2013) 2549-2582.
\bibitem{12}Jakub Konečný and Michal Hagara, One-Shot-Learning Gesture Recognition using HOG-HOF Features, Journal of Machine Learning Research 15 (2014) 2513-2532.
\bibitem{13}Feng Jiang, Multi-layered Hand Gesture Recognition with Kinect, Journal of Machine Learning Research 16 (2015) 227-254.
\bibitem{14}Jiaxiang Wu and Jian Cheng, Bayesian Co-Boosting for Multi-modal Gesture Recognition, Journal of Machine Learning Research 15 (2014) 3013-3036.
\bibitem{15}Nicholas Gillian and Joseph Paradiso, The Gesture Recognition Toolkit, Journal of Machine Learning Research 15 (2014) 3483-3487.
\bibitem{16}Long-Van Nguyen-Dinh, Robust Online Gesture Recognition with Crowdsourced Annotations, Journal of Machine Learning Research 15 (2014) 3187-3220.
\bibitem{17}Vassilis Pitsikalis, Athanasios Katsamanis, Stavros Theodorakis, and Petros Maragos, Multimodal Gesture Recognition via Multiple Hypotheses Rescoring, Journal of Machine Learning Research 16 (2015) 255-284. 
\bibitem{18}Sergio Escalera, Vassilis Athitsos, Isabelle Guyon, Challenges in multimodal gesture recognition, Journal of Machine Learning Research, vol. 17, pp. 1-54, 2016.
\bibitem{19}S. Escalera, J. Gonzàlez, X. Baró and J. Shotton, Guest Editors' Introduction to the Special Issue on Multimodal Human Pose Recovery and Behavior Analysis, IEEE Transactions on Pattern Analysis and Machine Intelligence (2016) 1489 - 1491.
\bibitem{20}F. Zhou and F. De la Torre, Spatio-Temporal Matching for Human Pose Estimation in Video, IEEE Transactions on Pattern Analysis and Machine Intelligence (2016) 1492 - 1504.
\bibitem{21}B. Wandt, H. Ackermann, and B. Rosenhahn, 3D Reconstruction of Human Motion from Monocular Image Sequences, IEEE Transactions on Pattern Analysis and Machine Intelligence (2016) 1505 - 1516.
\bibitem{22}M. Ye, Y. Shen, C. Du, Z. Pan, and R. Yang, Real-Time Simultaneous Pose and Shape Estimation for Articulated Objects Using a Single Depth Camera, IEEE Transactions on Pattern Analysis and Machine Intelligence (2016) 1517 - 1532.
\bibitem{23}T. von Marcard, G. Pons-Moli, and B. Rosenhahn, Human Pose Estimation from Video and IMUs, IEEE Transactions on Pattern Analysis and Machine Intelligence (2016) 1533 - 1547.
\bibitem{24}C.A. Corneanu, M.O. Simón, J.F. Cohn, and S.E. Guerrero, Survey on RGB, 3D, Thermal, and Multimodal Approaches for Facial Expression Recognition: History, Trends, and Affect-Related Applications, IEEE Transactions on Pattern Analysis and Machine Intelligence (2016) 1548 - 1568.
\bibitem{25}M. Sigalas, M. Pateraki, and P. Trahanias, Full-Body Pose Tracking - The Top View Reprojection Approach, IEEE Transactions on Pattern Analysis and Machine Intelligence (2016) 1569 - 1582.
\bibitem{26}D. Wu, L. Pigou, P.-J. Kindermans, N. D.-H. Le, L. Shao, J. Dambre, and J.-M. Odobez, Deep Dynamic Neural Networks for Multimodal Gesture Segmentation and Recognition, IEEE Transactions on Pattern Analysis and Machine Intelligence (2016) 1583 - 1597.
\bibitem{27}C.F. Crispim-Junior, V. Buso, K. Avgerinakis, G. Meditskos, A. Briassouli, J. Benois-Pineau, I. (Yiannis) Kompatsiaris, and F. Brémond, Semantic Event Fusion of Different Visual Modality Concepts for Activity Recognition, IEEE Transactions on Pattern Analysis and Machine Intelligence (2016) 1598 - 1611.
\bibitem{28}J.Y. Chang, Nonparametric Feature Matching Based Conditional Random Fields for Gesture Recognition from Multi-Modal Video, IEEE Transactions on Pattern Analysis and Machine Intelligence (2016) 1612 - 1625.
\bibitem{29}J. Wan, G. Guo, and S.Z. Li, Explore Efficient Local Features from RGB-D Data for One-Shot Learning Gesture Recognition, IEEE Transactions on Pattern Analysis and Machine Intelligence (2016) 1626 - 1639.
\bibitem{30}R. Zhao and A.M. Martinez, Labeled Graph Kernel for Behavior Analysis, IEEE Transactions on Pattern Analysis and Machine Intelligence (2016) 1640 - 1650.
\bibitem{31}M. Yu, L. Liu, and L. Shao, Structure-Preserving Binary Representations for RGB-D Action Recognition, IEEE Transactions on Pattern Analysis and Machine Intelligence (2016) 1651 - 1664.
\bibitem{32}Y. Panagakis, M.A. Nicolaou, S. Zafeiriou, and M. Pantic, Robust Correlated and Individual Component Analysis, IEEE Transactions on Pattern Analysis and Machine Intelligence (2016) 1665 - 1678.
\bibitem{33}K.K. Roudposhti, U. Nunes, and J, Dias, Probabilistic Social Behavior Analysis by Exploring Body Motion-Based Patterns, IEEE Transactions on Pattern Analysis and Machine Intelligence (2016) 1679 - 1691.
\bibitem{34}N. Neverova, C. Wolf, G. Taylor, and F. Nebout, ModDrop: Adaptative Multi-Modal Gesture Recognition, IEEE Transactions on Pattern Analysis and Machine Intelligence (2016) 1692 - 1706.
\bibitem{35}X. Alameda-Pineda, J. Staiano, R. Subramanian, L. Batrinca, E. Ricci, B. Lepri, O. Lanz, and N. Sebe, SALSA: A Novel Dataset for Multimodal Group Behavior Analysis, IEEE Transactions on Pattern Analysis and Machine Intelligence (2016) 1707 - 1720.
\bibitem{225}H. J.  Escalante, I.  Guyon, S.  Escalera, J.  Jaques Jr., X. Bar\'o, E. Viegas, Y. G\"u\'cl\"ut\"urk, U.G\"uçl\"u, M.  A. J. van Gerven, R. van Lier. Design of an Explainable Machine Learning Challenge for Video Interviews. Proceedings of IJCNN 2017.


\bibitem{44}Jiaxiang Wu, Jian Cheng, Chaoyang Zhao and Hanqing Lu, Fusing Multi-modal Features for Gesture Recognition, ICMI 2013.
\bibitem{45}Zhuowei Cai,Limin Wang, Xiaojiang Peng, Multi-view Super Vector for Action Recognition, ECCV, 2014.
\bibitem{46}Xiaojiang Peng, Limin Wang, Zhuowei Cai and Yu Qiao, Action and Gesture Temporal Spotting with Super Vector Representation, ECCV, 2014.
\bibitem{47}Natalia Neverova, Christian Wolf, Graham W. Taylor and Florian Nebout, Multi-scale deep learning for gesture detection and localization, ECCV, 2014.
\bibitem{48}Zhe Wang, Limin Wang, Wenbin Du, Qiao Yu, Exploring Fisher Vector and Deep Networks for Action Spotting, CVPR, 2015.
\bibitem{49}Limin Wang, Zhe Wang, Wenbin Du, Qiao Yu, Object-Scene Convolutional Neural Networks for Event Recognition in Images, CVPR, 2015.
\bibitem{50}Rasmus Rothe, Radu Timofte, Luc Van Gool, DEX: Deep EXpectation of apparent age from a single image, ICCV, 2015.
\bibitem{51}Mengyi Liu, Xin Liu, Yan Li, Xilin Chen, Alexander Hauptmann, Shiguang Shan, Exploiting Feature Hierarchies with Convolutional Neural Networks for Cultural Event Recognition, ICCV, 2015.
\bibitem{52}Grigory Antipov, Moez Baccouche, Sid-Ahmed Berrani, Jean-Luc Dugelay, Apparent Age Estimation from Face Images Combining General and Children-Specialized Deep Learning Models, CVPR, 2016.
\bibitem{53}Chenghua Li, Qi Kang, Guojing Ge, Qiang Song, Hanqing Lu, Jian Cheng, DeepBE: Learning Deep Binary Encoding for Multi-Label Classification, CVPR, 2016.
\bibitem{54}Kaipeng Zhang, Lianzhi Tan, Zhifeng Li, Yu Qiao, Gender and Smile Classification using Deep Convolutional Neural Networks, CVPR, 2016.
\bibitem{55}Chen-Lin Zhang, Hao Zhang, Xiu-Shen Wei and Jianxin Wu, Deep Bimodal Regression for Apparent Personality Analysis, ECCV 2016.
\bibitem{56}Albert Ali Salah, Furkan Gürpinar, Heysem Kaya, Combining Deep Facial and Ambient Features with Audio for First Impression Estimation, ICPR, 2016.
\bibitem{57}Yunan Li, Kuan Tian, Yingying Fan, Xin Xu, Rui Li, Video Gesture Recognition with RGB-D-S Data Based on 3D Convolutional Networks, ICPR, 2016.
\bibitem{58}Xiujuan Chai, Zhipeng Liu, Fang Yin, Zhuang Liu and Xilin Chen, Large-scale Continuous Gesture Recognition Using Convolutional Neural Networks, ICPR, 2016.



	
	
\end{thebibliography}



% \end{thebibliography}

\end{document}